\documentclass[runningheads]{llncs}

\usepackage[year=2026,ID=3482]{utils/eccv}

\usepackage{utils/eccvabbrv}

\usepackage{graphicx}
\usepackage{booktabs}
\usepackage{array}
\usepackage{multirow}
\usepackage[table]{xcolor}
\definecolor{salmon}{RGB}{180, 103, 97}    
\usepackage[accsupp]{axessibility}  

\usepackage{hyperref}

\usepackage{orcidlink}
\usepackage{wrapfig}
\usepackage{marvosym}

\begin{document}

\title{Evaluating and Enhancing Negation Comprehension in Remote Sensing MLLMs}

\titlerunning{RS-Neg \& NeFo}

\author{Haochen Han\inst{1}\thanks{Equal contribution. \noindent \textsuperscript{\Letter}Corresponding authors.} \and
Jue Wang\inst{2,1}\textsuperscript{*} \and
Alex Jinpeng Wang\inst{1,3}\textsuperscript{\Letter} \and
Fangming Liu\inst{1}\textsuperscript{\Letter}}

\authorrunning{H.~Han et al.}

\institute{Peng Cheng Laboratory, Shenzhen, China\\
\and
Tsinghua University, Beijing, China
\and
Central South University, Changsha, China\\
\href{https://hhc1997.github.io/RS-Neg-and-NeFo/}{\textcolor{eccvblue}{Code and Dataset: https://hhc1997.github.io/RS-Neg-and-NeFo/}}}

\maketitle

\begin{abstract}
Multimodal Large Language Models (MLLMs) have demonstrated remarkable success in various Remote Sensing (RS) tasks. However, their ability to comprehend negation remains underexplored, limiting deployment in real-world applications where models must explicitly identify what is false or absent, e.g., emergency responders need to locate non-flooded routes for evacuation. To comprehensively study this limitation, we introduce RS-Neg, the first benchmark to evaluate negation understanding across region-level to scene-level tasks. Specifically, we design an automated data generation pipeline for RS imagery, using LLMs to synthesize diverse negation queries, and introduce a dynamic visual focus module for verification. Our evaluation reveals that advanced RS MLLMs struggle with negation, exhibiting hallucinations and substantial performance degradation. To close this gap, we propose NeFo, a novel test-time adaptation method that explicitly incorporates the logical role of negation into the model optimization. Remarkably, using about 5\% unlabeled test samples, NeFo significantly improves the negation understanding of models and shows strong generalization to unseen tasks.

  \keywords{Multimodal \and Negation Understanding \and Remote Sensing}
\end{abstract}

\section{Introduction}
\label{sec:intro}

Negation is a sine qua non of every linguistic system, helping humans describe their perceived world by specifying what is absent, false, or contrary to observation. Studies in cognitive science \cite{hasson2006does} reveal that negation understanding emerges before broader world knowledge develops: 18-month-olds can already interpret novel objects with negative sentences. 

However, such basic reasoning ability remains underdeveloped in advanced AI systems---Multimodal Large Language Models (MLLM). Recent research \cite{zhang2025negvqa} shows that even MLLM with 70 billion parameters suffer significant performance degradation when handling negation. This limitation becomes especially critical in RS vision-language tasks due to the inevitable demand for negation understanding. For example, models need to answer `how many buildings are not flooded' in natural disaster monitoring, or locate `routes without ice coverage' in transportation planning. Undoubtedly, misunderstanding negation could cause targets contrary to user intent and pose serious safety risks.

\begin{wrapfigure}{r}{0.55\textwidth}
    \begin{minipage}{0.55\textwidth}
        \centering  
        \vspace{-1mm}
        \scalebox{1.00}
        {
            \includegraphics[width=\textwidth]{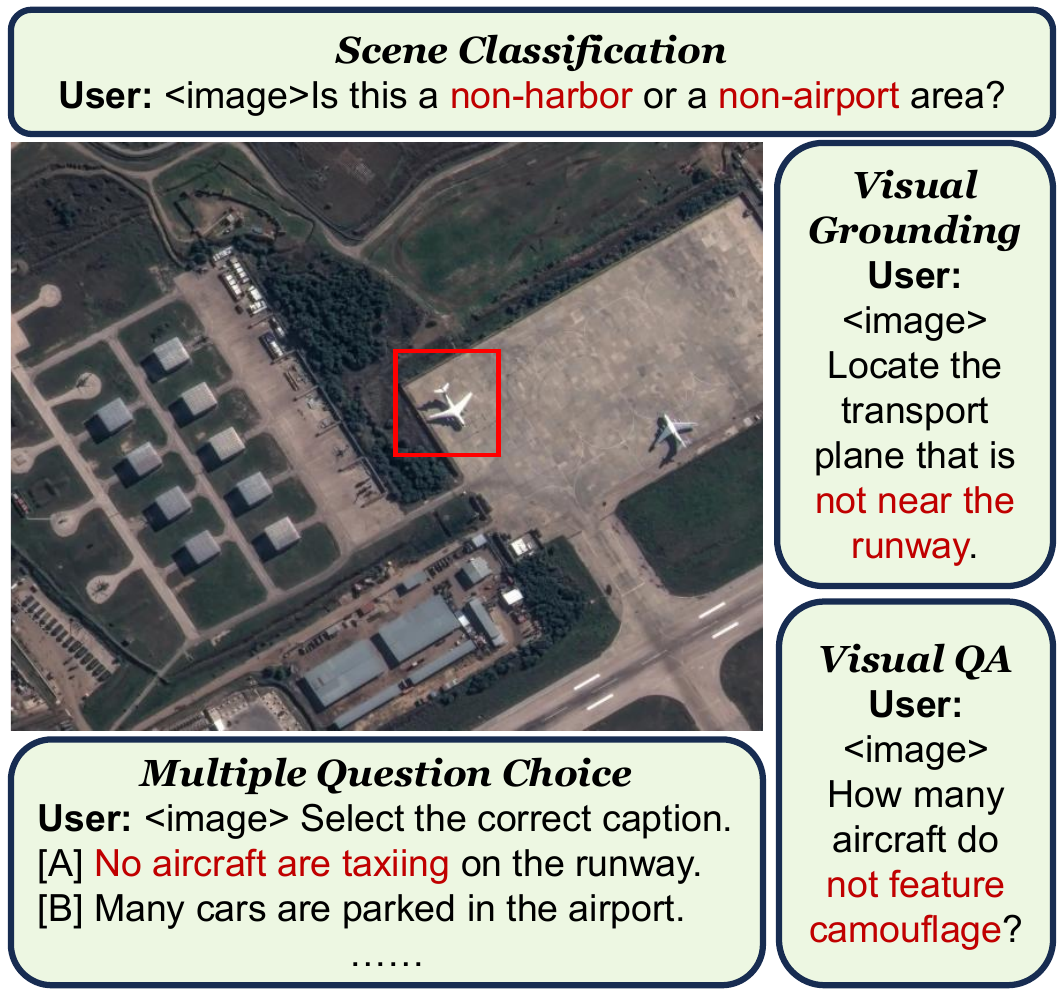}  
        }
        \captionof{figure}{A toy example to illustrate our negation understanding tasks for remote sensing.}
        \label{fig:toy_example}
        \vspace{-2mm}
    \end{minipage}
\end{wrapfigure}

Recently, several attempts have been proposed to assess how well MLLMs handle negation. NegVQA \cite{zhang2025negvqa} constructs two-choice questions covering different negation scenarios and image distributions. GaslightingBench \cite{zhu2025calling} introduces multiple-choice questions to evaluate models' resistance to misleading negation arguments while maintaining logical consistency. In addition, some studies \cite{singh2024learn,alhamoud2025vision} design negation-enriched retrieval systems for testing embedding models like CLIP \cite{radford2021learning}. Despite the success, these benchmarks focus on general-domain vision-language tasks, and the proposed data generation pipelines do not apply to the fine-grained RS imagery \cite{li2025simple,li2026co}. This raises two practical but untouched research questions: (1) How to comprehensively evaluate the negation understanding capability of MLLMs in RS scenarios? (2) How can we enhance the robustness of MLLMs when handling negation queries?

To address the first question, we propose RS-Neg, a benchmark designed to evaluate negation understanding in RS. As illustrated in \cref{fig:toy_example}, RS-Neg comprises 22K samples covering four fundamental tasks that span region-level to scene-level scenarios. To construct RS-Neg, we leverage LLMs to synthesize natural and diverse negation queries from existing RS datasets, and introduce a dynamic visual focus procedure to ensure the generated negation concepts are absent from the visual content. Through our evaluation framework, we uncover that RS MLLMs face significant challenges when deployed in negation-conditioned environments.

To address the second question, we propose NeFo, an efficient Test-Time Adaptation (TTA) method that steers MLLMs to focus on negation-related semantics using only unlabeled test data. Specifically, we explicitly incorporate the logical role of negation into model optimization that maximizes the output discrepancy between the negated query and its negation-masked variant. In addition, to prevent catastrophic forgetting, we employ the original model as a teacher to regularize the predictions on affirmative queries.

The main contributions of our work are summarized as follows:

\begin{itemize}

\item[1.] We introduce RS-Neg, the first benchmark for evaluating negation understanding in RS MLLMs. RS-Neg comprises 22K samples spanning region-level to scene-level scenarios, constructed through a meticulous pipeline that combines LLM-driven query synthesis with dynamic visual verification.

\item[2.] We propose NeFo, the first test-time adaptation method to enhance negation understanding in MLLMs. Without relying on additional annotations or teacher models, NeFo incorporates the logical role of negation into the optimization objective in a self-supervised manner.

\item[3.] Extensive experiments verify the effectiveness of the proposed method. With only 2.5 million training parameters and minimal unlabeled samples, NeFo achieves significant gains across different base MLLMs while demonstrating strong generalization to unseen tasks.
\end{itemize}

\section{Related Works}
\subsection{Evaluating Negation Understanding in VLMs.}
Vision-language models learn rich knowledge from large-scale multimodal data, demonstrating strong capabilities across diverse domains like RS \cite{guo2024skysense,nedungadi2024mmearth,zhu2025skysense,zhang2025skysense,danish2025geobench}. However, recent studies have highlighted that such powerful models struggle to understand negation—a sine qua non of every human language. Based on different VLM architectures, current works can be roughly divided into two categories: 1) Embedding VLMS. To evaluate the performance of embedding models (e.g., CLIP \cite{radford2021learning} and ALIGN \cite{jia2021scaling}) in negation scenarios, recent works CREPE \cite{ma2023crepe} and CC-Neg \cite{singh2024learn} proposed template-based image-caption benchmarks for compositional understanding with negation. To improve the linguistic diversity of negated caption, NegBench \cite{alhamoud2025vision} resort to LLMs to generate more natural negated captions, spanning images and videos. Evaluations on these benchmarks show that even billion-scale models like CLIP, struggles with negation in retrieval, generation, and segmentation tasks. 2) Generative VLMS. Recent studies \cite{zhu2025calling} show that advanced MLLMs such as GPT-4o and Claude-3.5 can generate hallucinations when handling negation arguments from users. To evaluate this limitation in MLLMs, NegVQA \cite{zhang2025negvqa} construct two-choice questions covering diverse negation scenarios, showing significant performance drops across 20 popular MLLMs (e.g., Qwen2-VL \cite{wang2024qwen2} and DeepSeek-VL \cite{lu2024deepseek}).

Unlike prior works that mainly provide diagnostic tools for general-domain web images, we focus on negation understanding in RS tasks. Our proposed data pipeline can handle small targets in RS images, generating comprehensive evaluation across region and scene levels.

\subsection{Enhancing Negation Understanding in VLMs.}
Recent methods have explored improving the negation understanding abilities in vision-language embedding models, which generally follow two research lines: 1) Post-training. Several data-driven methods attempted to fine-tune VLMs on massive negated image-text pairs. For example, NegCLIP \cite{yuksekgonul2022and} creates negative captions via swapped relations to enhance compositional reasoning. CoNCLIP \cite{singh2024learn} integrates template-based negation samples into the contrastive loss. The recent advances use LLMs and MLLMs \cite{alhamoud2025vision,park2025know} to synthesize million-scale negation samples for post-training. 2) Test-time adaptation method. In contrast to data-driven methods that require massive training samples, the recent study NEAT \cite{han2025negation} argues that the key to negation understanding is a distribution shift problem. It proposes a TTA method to correct the semantic consistency.

Despite the success, these methods target realigning vision-language embeddings and are difficult to extend to MLLMs. As collecting massive labeled RS data is expensive, we focus on TTA to update MLLMs during test time with only unlabeled test data. Given the inherent demand for negation queries in RS scenarios, we believe this study could provide some practical insights to real-world tasks such as disaster response.

\section{RS-Neg: Benchmark for Negation in Remote Sensing}
\label{sec:RemoteNeg}
We propose RS-Neg, a multi-level benchmark for assessing negation comprehension in RS MLLMs, which consists of four core tasks spanning from region-level to scene-level evaluation. Specifically, for visual conversation tasks such as visual question answering, multiple choice questions, and visual grounding, we design an automated pipeline using LLMs and MLLMs to synthesize high-quality negation queries. A dynamic visual focus procedure is further introduced to localize fine-grained objects in RS images. For tasks with structured labels, e.g., scene classification, we apply rule-based reformulation to incorporate negated labels as distractors. 


\subsection{RS Negation Data Pipeline}
Given the high cost of RS annotation, our pipeline builds upon widely used image-caption datasets, readily transforming affirmative descriptions into negation-aware counterparts. Benefiting from the powerful capabilities of LLMs or MLLMs, prior works \cite{singh2024learn,park2025know,alhamoud2025vision,zhang2025negvqa} have explored negated data augmentation through in-context learning. However, these methods focus on general-domain web images and face two key obstacles when handling RS imagery. 

First, these studies mainly generate object-level negation by adding absent entities to the caption, e.g., `a dog with no leash', while RS scenarios often demand property-level reasoning. For example, models might respond `find a road not covered in ice' or `how many non-flooded buildings' in disaster assistance. Second, targets of interest in RS images are typically small-scale (less than 32$\times$32 pixels \cite{zhang2024FFCA}) relative to the large field of view, posing significant challenges for prior pipelines with global image perception. 

\begin{figure}[t]
\centering  
\includegraphics[width=\columnwidth]{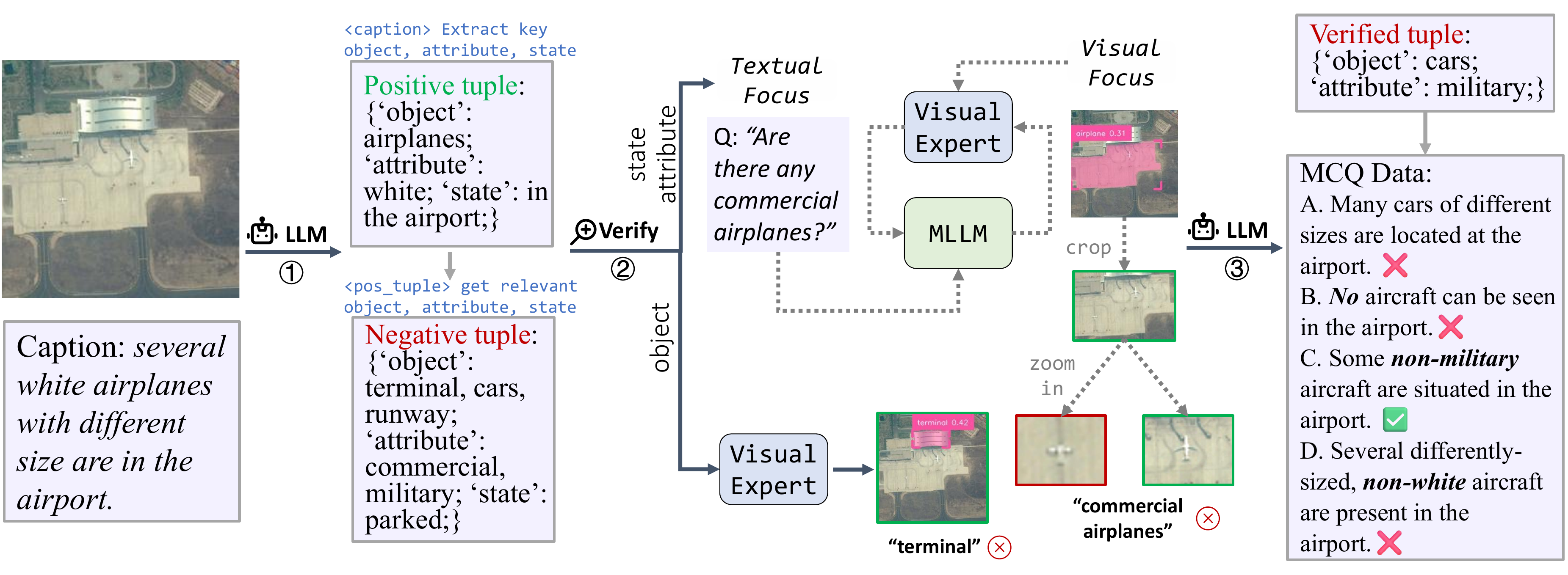}
\caption{Pipeline for constructing RS-Neg dataset, with MCQ as an example. Given an RS image-caption pair, we first use an LLM to extract concepts from the caption and generate corresponding negative counterparts. Second, we employ an MCTS-based visual focus method to verify these negative concepts against the image. Third, we use an LLM to formulate task-specific samples and enhance linguistic diversity.}
\label{fig:pipeline}
\end{figure}

To address such limitations, our pipeline first uses LLMs to generate negation-conditioned captions at object, attribute, and state levels, enabling more realistic and diverse patterns. We then employ the dynamic visual focus method \cite{li2025dyfo} to verify whether generated negation concepts are correct, where MLLMs and visual expert models interact via Monte Carlo Tree Search (MCTS) to focus on key visual regions. Finally, these filtered negation captions are transformed into task-specific visual conversation data via LLM augmentation. The overall pipeline proceeds as follows (Detailed steps are presented in Supplementary A.):

\textbf{Extracting Candidate Negated Concepts.} Given a RS image-caption pair, we first parse the caption with the LLM to extract the positive tuple, i.e., mentioned objects, corresponding attributes (e.g., color or shape), and states (e.g., action or condition). The extracted tuple is then fed to the LLM to generate the candidate negated tuple—plausible elements related to the context but absent from the caption.

\textbf{Verifying Negated Concepts.} Since the above process considers only text input, we then verify whether these generated negative concepts are present in the image. Although MLLMs \cite{dai2023instructblip,liu2023visual} can accurately identify the absence of content in web images, they struggle to handle RS images due to small-scale targets. As shown in Supplementary B, even the powerful closed-source MLLM (Claude Opus 4.5) may fail in such scenarios. To address this, we adopt Dynamic Focus (DyFo) \cite{li2025dyfo}, a training-free MCTS-based visual search method to amplify key visual information. Specifically, for object-level elements, we directly query visual experts (e.g., Grounding Dino \cite{liu2024grounding}) for presence detection. For the attribute or state element, we combine it with the corresponding object to form the root node input for tree search. In the expansion phase, new child nodes are generated by sampling from two action spaces: 1) Semantic Focus. The visual expert localizes this target and crops the specific region, yielding a smaller focused image. 2) Spatial Zoom In. As shown in \cref{fig:pipeline}, the cropped region may contain numerous instances of the target object (e.g., roof) with diverse properties (e.g., red and orange). We magnify the region by a scale factor, progressively narrowing down to enable instance-level inspection. The MLLM  (Qwen2.5-VL-7B) then verifies each node and provides reward signals to guide the search. Details for this process is provided in the supplementary C.

\textbf{Task-Specific Data Generation.}
With the filtered absent elements, we formulate negation data tailored to different tasks: 1) VQA. We construct binary questions by negating either present or absent elements. For example, the target is modified as, ``Does this image show \{pos\_object\} without \{neg\_object\}? Yes.'', ``Is there any \{pos\_object\} not \{pos\_state\} seen? No.'', or ``Is there a non-\{neg\_attribute\} \{pos\_object\} seen? Yes.'' All questions are further paraphrased by LLMs for linguistic diversity. 2) MCQ. For each image, we generate one correct caption and three distractors, with at least one containing negation. The MCQ task tests the model's ability to distinguish subtle differences among affirmative, negated, and hybrid descriptions, reflecting more realistic decision-making. All MCQs are also rewritten by LLMs for linguistic diversity. 3) Visual Grounding. For captions with bounding box annotations, we modify captions by negating absent elements while referring to the same target object, thus preserving the ground truth localization. 4) Scene Classification. Beyond the LLM-based pipeline above, we also introduce a simple rule-based reformulation that incorporates negated labels as distractors for the classification task. Examples of each task in RS-Neg are provided in the supplementary D.

\subsection{Dataset and Evaluation}

\begin{wrapfigure}{r}{0.55\textwidth}
    \begin{minipage}{0.55\textwidth}
        \centering  
        \vspace{-6mm}
        \captionof{table}{RS-Neg dataset statistics.}
        \label{tab:data_statistics}
        \setlength{\tabcolsep}{3pt}
        \scalebox{0.85}
        {
            \begin{tabular}{lcccc}
            \toprule
            Task  & Object & Attribute & State & Total \\
            \midrule
            VQA  &5,675       &2,379       &1,151       &9,205  \\
            MCQ   &4,502      &791      & 382      & 5,675  \\
            Grounding &1,494     & 671     & 319     & 2,484  \\
            Classification & -       & -      & -       & 5,100  \\
            \bottomrule
            \end{tabular}%
        }
        \vspace{-2mm}
    \end{minipage}
\end{wrapfigure}

\textbf{Dataset Overview.} Using the proposed pipeline, we construct the RS-Neg dataset from 7 popular RS datasets, yielding 9,205 VQA samples and 5,675 MCQ samples sourced from NWPU Caption \cite{cheng2022nwpu}, RSICD \cite{lu2017exploring}, Sydney Caption \cite{qu2016deep}, and VRSBench \cite{li2024vrsbench}, 2,484 visual grounding samples from VRSBench \cite{li2024vrsbench}, and 5,100 scene classification samples from AID \cite{xia2017aid} and UCMerced \cite{yang2010bag}. In total, RS-Neg contains 22,464 samples spanning four fundamental RS tasks, providing a comprehensive evaluation of negation understanding from region-level to scene-level scenarios.


\begin{figure}[t]
    \centering
    \begin{subfigure}[b]{0.49\textwidth}
        \centering
        \includegraphics[width=\textwidth]{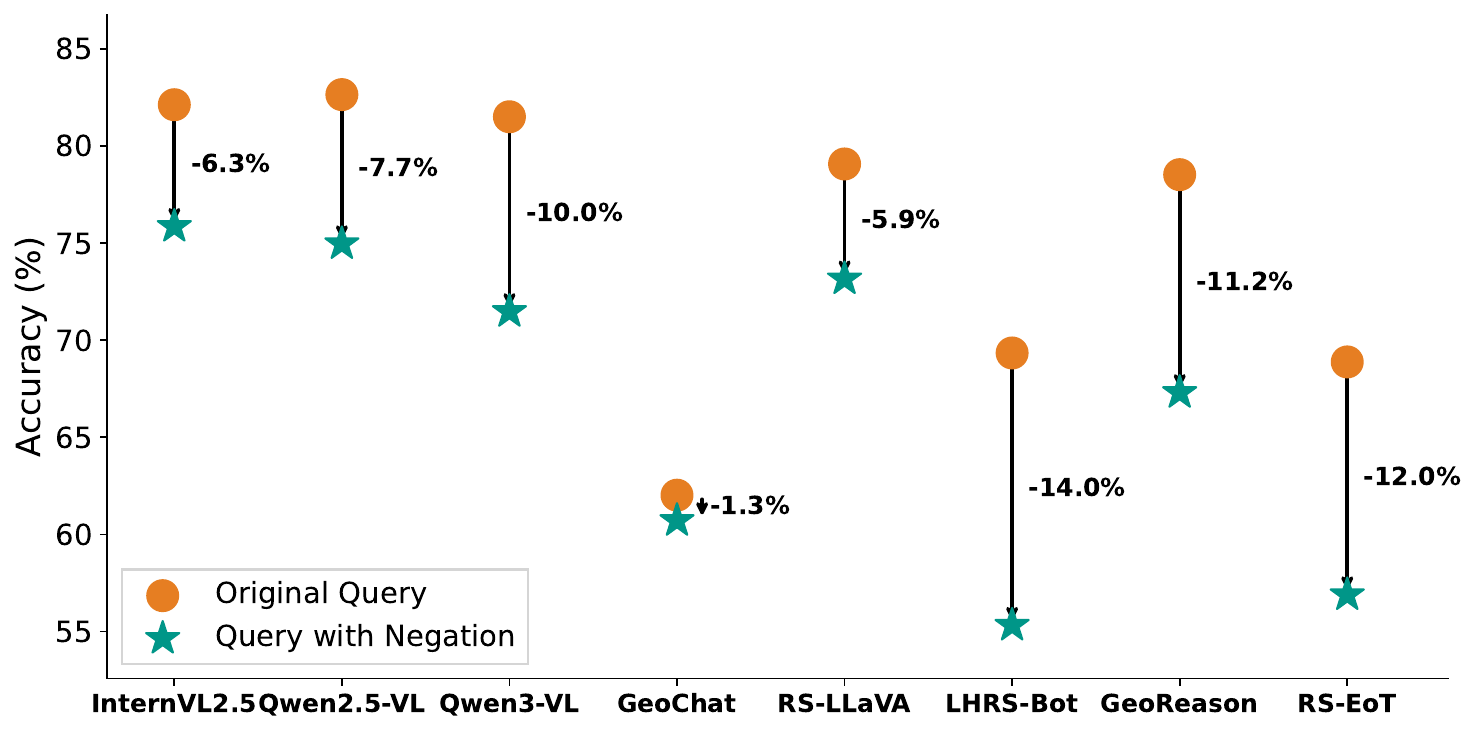}
        \caption{Visual Question Answering}
        \label{fig:neg_vqa}
    \end{subfigure}
    \hfill
    \begin{subfigure}[b]{0.49\textwidth}
        \centering
        \includegraphics[width=\textwidth]{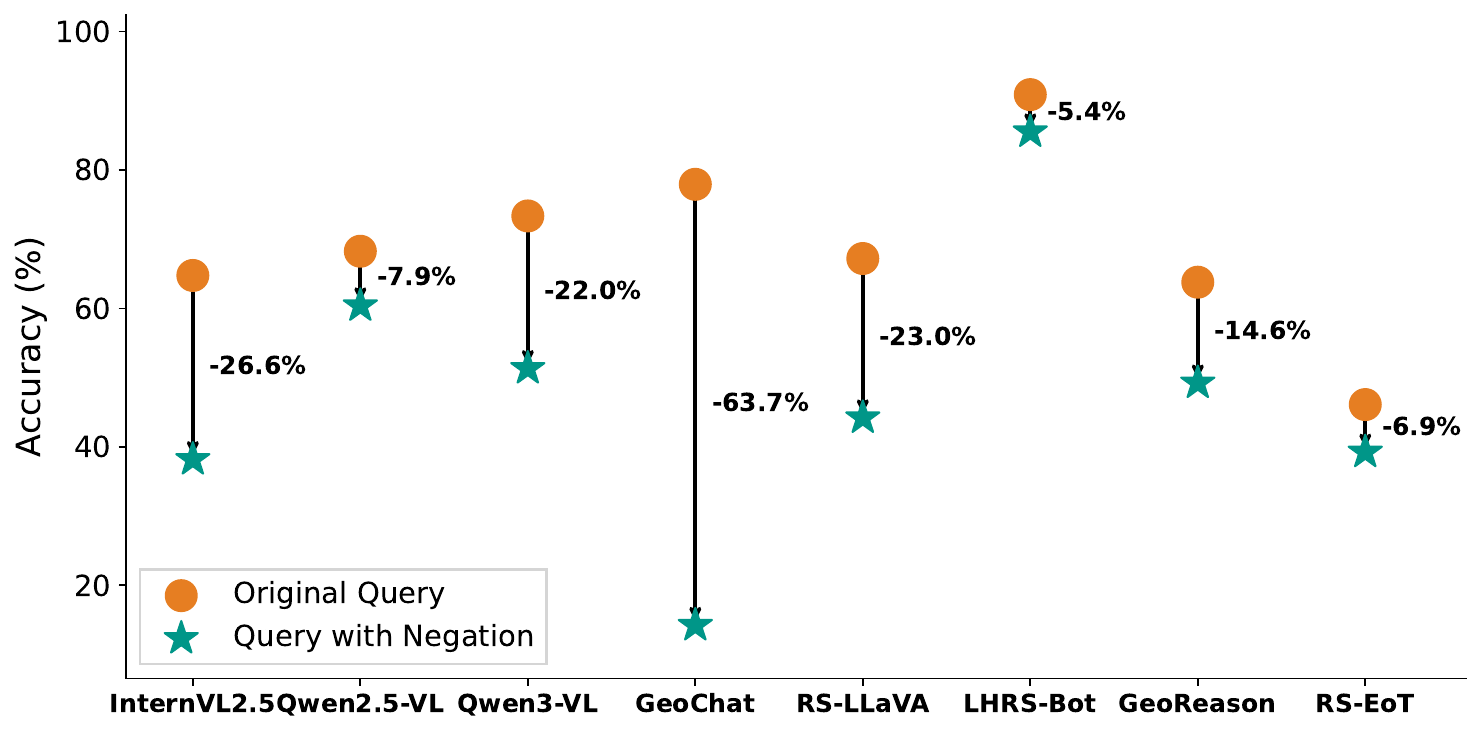}
        \caption{Scene Classification}
        \label{fig:neg_cls}
    \end{subfigure}
    
    
    \begin{subfigure}[b]{0.49\textwidth}
        \centering
        \includegraphics[width=\textwidth]{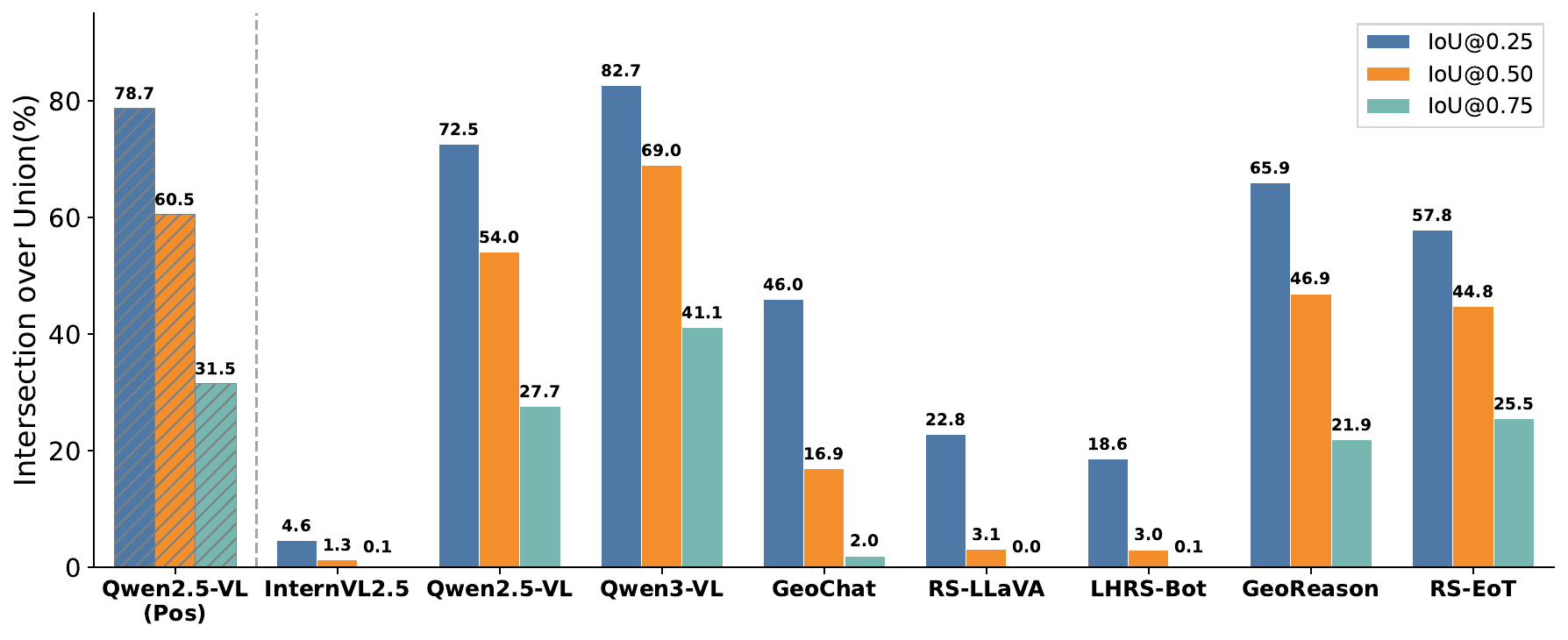}
        \caption{Visual Grounding}
        \label{fig:neg_vrs}
    \end{subfigure}
    \hfill
    \begin{subfigure}[b]{0.49\textwidth}
        \centering
        \includegraphics[width=\textwidth]{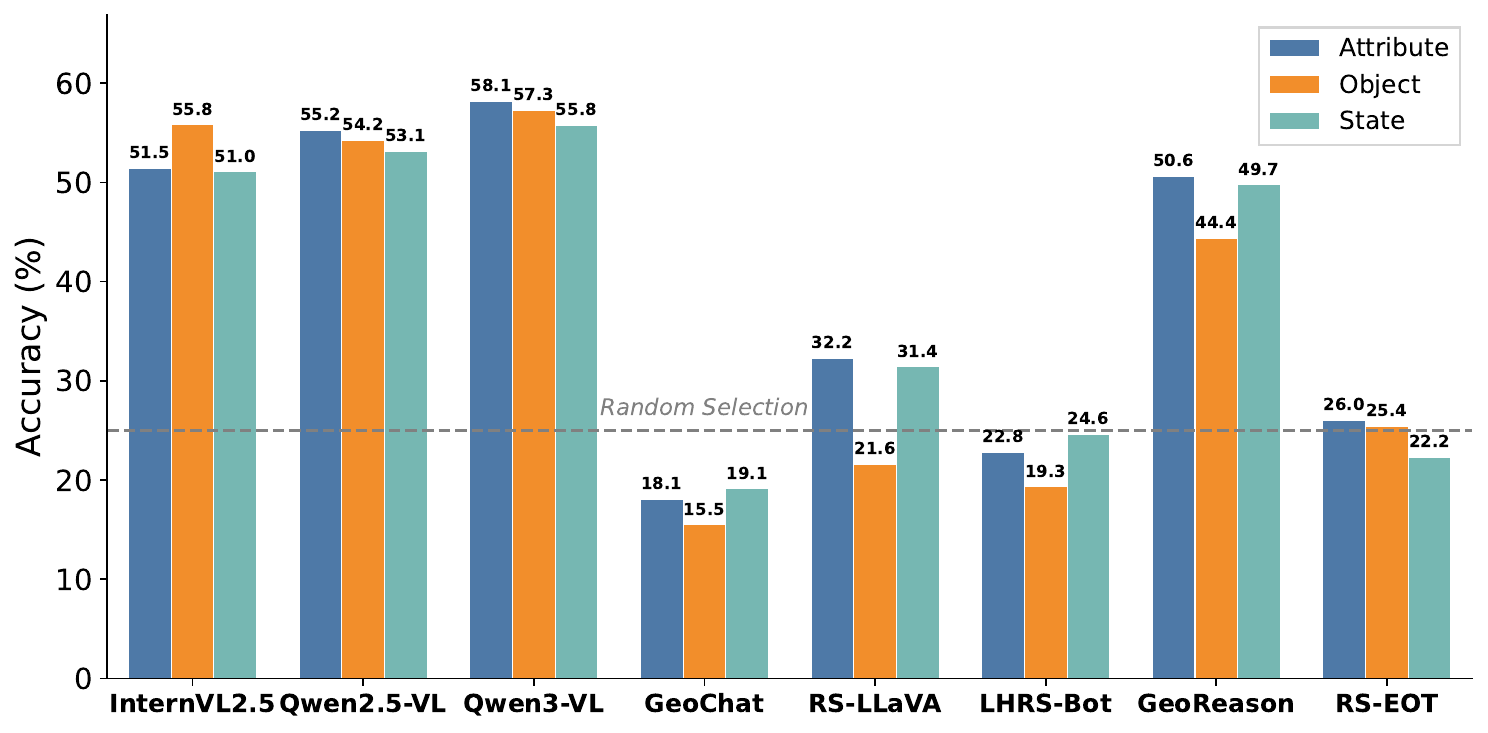}
        \caption{Multiple Choice Questions}
        \label{fig:gap_mcq}
    \end{subfigure}
    \caption{Model performance on RS-Neg across different MLLMs and tasks. (a) and (b) show the performance drop compared to original queries on VQA and classification tasks, respectively. (c) reports the results on the visual grounding task, where we include the performance of Qwen2.5-VL on corresponding affirmative queries as a positive baseline. (d) reports the results on the MCQ task, where several RS-specific models exhibit notably poor performance, falling below the random guessing baseline.} 
    \label{fig:gap_comparison}
\end{figure}

\textbf{RS MLLMs Struggle with Negated Queries.}
In this section, we benchmark the negation abilities of different MLLMs using our RS-Neg, including general-purpose, RS-specific, and reasoning-augmented models. For fair comparison, we selected 8 top-performing models with similar parameter scales, i.e., Qwen2.5-VL-7B-Instruct \cite{bai2025qwen2}, InternVL2.5-8B \cite{chen2024expanding}, Qwen3-VL-7B-Instruct, GeoChat-7B \cite{kuckreja2024geochat}, RS-LLaVA-7B \cite{bazi2024rs}, LHRS-Bot-7B \cite{muhtar2024lhrs}, GeoReason-7B \cite{li2026georeason}, and RS-EOT-7B \cite{shao2025asking}. Note that GeoReason and RS-EOT are strong reasoning MLLMs designed to mitigate logical hallucinations in RS tasks.

Our evaluation reveals that current MLLMs consistently underperform on negation queries compared to their affirmative counterparts. As shown in \cref{fig:neg_vqa} and \cref{fig:neg_cls}, all models suffer significant performance drops on scene-level negated questions, with average degradation of 8.6\% on VQA task and 21.3\% on classification task. Notably, despite GeoChat's strong performance on original scene classification, its accuracy drops by 63.7\% when handling negation, indicating severe overfitting to the query pattern. This limitation also holds at region-level tasks. \cref{fig:neg_vrs} shows the visual grounding results, where we adopt the representative Qwen2.5-VL on original queries as a positive baseline given the substantial performance variance across models. Specifically, Qwen2.5-VL loses 6.2\%, 6.5\%, and 3.8\% at IoU@0.25, IoU@0.5, and IoU@0.75, respectively. While for the more challenging MCQ task, which requires models to parse subtle yet critical differences between affirmative and negated captions, most RS-specific models (e.g., GeoChat, RS-LLaVA, and LHRS-Bot) even fall below the random guessing baseline. In addition, reasoning-augmented models such as GeoReason and RS-EOT offer no advantage, suggesting that chain-of-thought reasoning does not address the fundamental deficiency in negation understanding.

\section{Method}
\label{sec:method}
\begin{figure}[t]
\centering  
\includegraphics[width=\columnwidth]{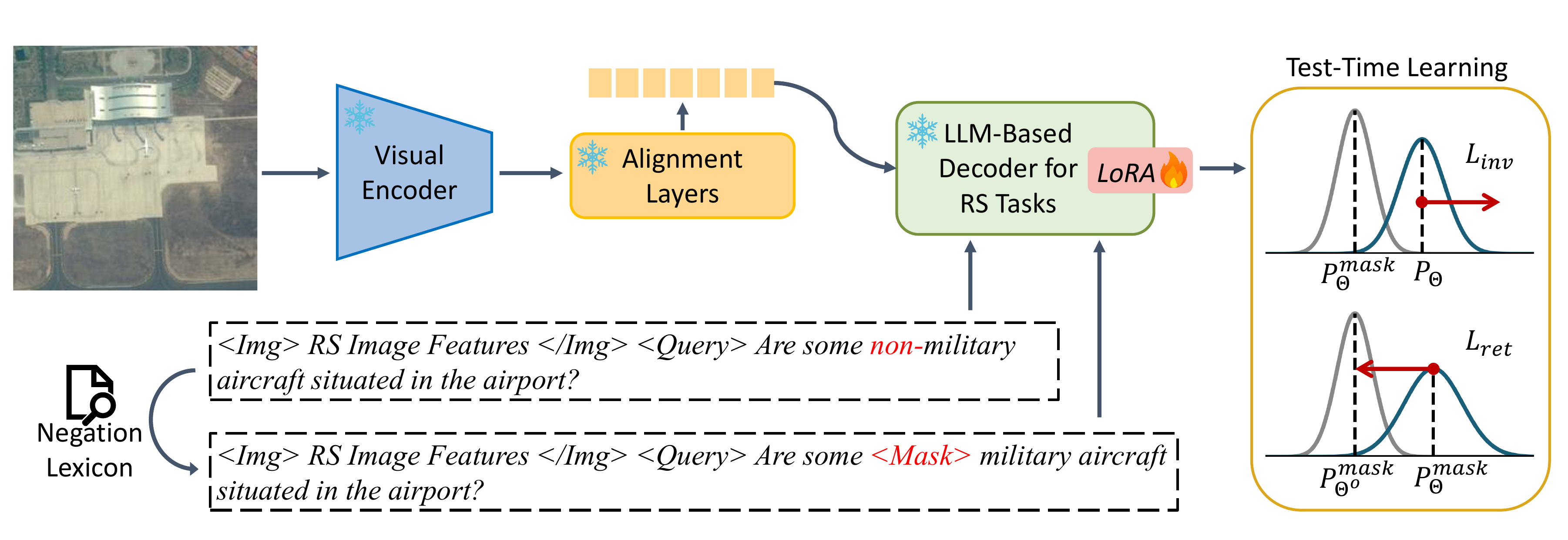}
\caption{An overview of the proposed NeFo. Given the test sample with a negation query, we first use a lexicon to construct the negation-masked text counterpart. Then we use LoRA to fine-tune MLLMs with two training objectives: the Truth-Value Inversion Loss that forces the model to focus on negation-related information, and the Knowledge Retaining Loss that preserves the foundational knowledge on negation-masked variants.}
\label{fig:method}
\end{figure}

To bridge this gap, we propose NeFo, a test-time adaptation method that guides the model to focus on negation-related semantics. The key idea behind NeFo is to explicitly incorporate the logical role \cite{horn2001natural} of negation into model optimization—a unary operator that inverts the truth-value of a statement. As shown in \cref{fig:method}, without relying on additional annotations or teacher models, NeFo enhances negation understanding in a fully self-supervised manner and enables efficient adaptation through lightweight LoRA updates. We next describe the training objectives of NeFo in detail.

\subsection{Problem Formulation}
Without loss of generality, let $f_{\Theta^o}(\cdot)$ denote a trained Multimodal Large Language Model (MLLM) that can generate visually grounded responses for RS tasks. Although trained on massive data, text instructions are mainly expressed
affirmatively, hindering model generalization for negation understanding. During real-world deployments, given a set of test samples $\{(v_i, q_i)\}_{i=1}^N$, where $v_i$ is an RS image and $q_i$ is the corresponding user query that may contain negation, the model can generate hallucinated outputs $f_{\Theta^o}(y|v_i, q_i)$ due to the poor ability of negation understanding.

\subsection{The Training Objective}
\textbf{Truth-Value Inversion.} From a linguistic-logical perspective\cite{horn2001natural,deprez2020oxford}, negation acts as an operator toggling the truth value of statements between true and false. Ideally, a well-behaved MLLM should also adhere to this logical principle in its reasoning, e.g., generating totally opposite responses to `Is the area flooded?' versus `Is the area not flooded?'. Based on this intuition, we propose a truth-value inversion loss that maximizes the output discrepancy between the negation-conditioned query and its negation-masked variant. Specifically, given the query $q$ containing negation, we first construct $q^{mask}$ by masking out the negation tokens. As negation tokens in natural language are limited (e.g., no, not, and without), we use a lexicon-based filtering as our masking strategy. Formally, let $\mathcal{P}_{\Theta}(y|v,q^{mask})$ denote the probability of generating output $y$ conditioned on the affirmative input. We enforce distinct output of the negated counterpart by maximizing the KL divergence:

\begin{equation}\label{eq: truth inversion}
\mathcal{L}_{inv} =
 - \text{max}(\mathbb{D}_{\text{KL}}[\mathcal{P}_{\Theta}(y|v, q) \| \mathcal{P}_{\Theta}(y|v, q^{mask})], \alpha ),
\end{equation}
where $\alpha$ is a clamp parameter to ensure training stability. Equation \eqref{eq: truth inversion} can amplify output differences arising from subtle input changes, which encourages the model to focus on key negation-related information.

\textbf{Knowledge Retaining.} To prevent catastrophic forgetting during TTA, we further employ the original MLLM as a teacher to regularize the model's predictions. Since the original model may produce noisy responses to negated queries, we only preserve foundational knowledge learned on negation-masked variants. Meanwhile, we introduce a standard entropy minimization as an auxiliary objective to better adapt the MLLM to the test data. Formally, the knowledge retaining loss is defined as follows:
\begin{equation}\label{eq: knowledge retaining}
\mathcal{L}_{ret} =
 \beta\mathbb{D}_{\text{KL}}[\mathcal{P}_{\Theta^o}(y|v, q^{mask}) \| \mathcal{P}_{\Theta}(y|v, q^{mask})] + \gamma\mathcal{H}(\mathcal{P}_{\Theta}(y|v, q^{mask})),
\end{equation}
where $\mathcal{H}(\cdot)$ denote the entropy term, $\beta$ and $\gamma$ are balance factors.

\textbf{TTA Strategy.} Combining the knowledge retaining loss with truth-value inversion loss can provide clear optimization guidance for negated queries. Thus, the final TTA objective is defined as:
\begin{equation}\label{eq: total_loss}
\min_{\tilde{\Theta}}\mathcal{L} = \mathcal{L}_{inv} +\mathcal{L}_{ret},
\end{equation}
where $\tilde{\Theta}\subseteq \Theta$ denotes the subset of model parameters to
be updated. We adopt Low-Rank Adaptation to fine-tune the attention layers in the LLM module, enabling parameter-efficient training for real-world deployment.


\section{Experiments}
\label{sec:Experiments}
In this section, we verify the effectiveness of NeFo by addressing the following key questions: (1) Test-Time Efficacy: Does NeFo successfully enhance MLLM robustness to negation during inference? (2) Generalization Capability: Can NeFo learn generalizable negation patterns rather than simply fitting to the test data?

\subsection{Implementation Details}
We conduct the test-time training experiments on two negation image-caption datasets in RS-Neg, i.e., VQA and MCQ. Note that for MCQ data, we prompt each option as "Does this caption describe the image? \texttt{<image><caption>}", and only include samples with negation tokens for training. As real-world deployment requires models to adapt quickly with limited data, we perform online learning using only minimal test samples and report the performance on the full  test set. Specifically, for all experiments, we use $150$ and $300$ samples from VQA and MCQ data for TTA, respectively. As NeFo is a general framework compatible with most MLLMs, we select four representative models on RS tasks as source models for evaluation, i.e., Qwen2.5-VL, Qwen3-VL, RS-LLaVA, and GeoReason. We use AdamW as the update with a batch size of 1. The parameters $\alpha$ and $\beta$ are set to $0.4$ and $0.4$, respectively. For the LoRA setup, we set the rank to 8 and the scaling factor to 16. The learning rate and $\gamma$ settings for each model are presented in supplementary E.

\subsection{Comparison Experiments}

In this section, we compare our method with three SOTA test-time adaptation methods, i.e., TENT \cite{wang2020tent}, SAR \cite{niutowards}, and TLM \cite{pmlr-v267-hu25z}, under negation queries. We study both the effectiveness and generalization capability of NeFo by conducting the following analysis experiments.

\begin{table*}[t]
  \centering
  \setlength{\tabcolsep}{4pt}
    \setlength{\aboverulesep}{0pt}
\setlength{\belowrulesep}{0pt}
\setlength{\extrarowheight}{0pt}
  \renewcommand\arraystretch{0.86}
      \caption{Comparisons with state-of-the-art TTL methods on RS-Neg VQA and RS-Neg MCQ datasets under negated queries. The best results are marked in \textbf{bold}.}
    \begin{tabular}{l|ccc|c|ccc|c}
    \toprule
    \multirow{1}[4]{*}{Method} & \multicolumn{4}{c|}{RS-Neg VQA} & \multicolumn{4}{c}{RS-Neg MCQ} \\
\cmidrule{2-9}          & object & attribute & state & total & object & attribute & state & total \\
    \midrule
    Qwen2.5-VL & 80.19 & 69.48 & 60.47 & 74.96 & 54.22 & 55.25 & 53.14 & 54.29 \\
    ·Tent & 77.52 & 68.89 & 59.95 & 73.09 & 56.11 & 55.63 & 52.00 & 55.77 \\
    ·SAR  & 68.35 & 61.75 & 56.91 & 65.21 & 55.84 & 55.37 & 52.09 & 55.52 \\
    ·TLM  & 72.67 & 66.20 & 59.08 & 69.30 & 49.42 & 48.42 & 46.86 & 49.43 \\
    \rowcolor{salmon!20} ·NeFo & \textbf{86.26} & \textbf{72.09} & \textbf{61.69} & \textbf{79.52} & \textbf{65.73} & \textbf{65.49} & \textbf{62.30} & \textbf{65.46} \\
    \midrule
    Qwen3-VL & 76.00 & 63.64 & 65.33 & 71.47 & 57.29 & 58.15 & 55.76 & 57.30 \\
    ·Tent & 76.72 & 64.69 & 67.07 & 72.41 & 61.11 & 61.82 & 58.38 & 61.02 \\
    ·SAR  & 79.79 & 65.57 & 67.42 & 74.57 & 61.62 & 62.58 & 59.95 & 61.64 \\
    ·TLM  & 67.24 & 62.51 & \textbf{70.03} & 66.37 & 58.09 & 58.66 & 53.66 & 57.87 \\
    \rowcolor{salmon!20} ·NeFo & \textbf{81.04} & \textbf{66.08} & 66.99 & \textbf{75.42} & \textbf{64.90} & \textbf{63.84} & \textbf{61.78} & \textbf{64.55} \\
    \midrule
    RS-LLaVA & 77.32 & 65.07 & 69.24 & 73.15 & 21.57 & 32.24 & 31.41 & 23.72 \\
    ·Tent & 77.48 & 67.17 & 68.38 & 73.68 & 19.39 & 30.72 & 30.37 & 21.71 \\
    ·SAR  & 72.09 & 60.99 & 67.33 & 68.63 & 20.19 & 31.23 & 31.41 & 22.48 \\
    ·TLM  & 76.25 & \textbf{66.58} & 67.51 & 72.66 & 20.64 & 31.73 & 30.63 & 22.85 \\
    \rowcolor{salmon!20} ·NeFo & \textbf{79.05} & 65.62 & \textbf{72.46} & \textbf{74.75} & \textbf{22.41} & \textbf{32.49} & \textbf{31.94} & \textbf{24.46} \\
    \midrule
    GeoReason & 72.26 & 57.38 & \textbf{63.34} & 67.30 & 44.36 & 50.57 & 49.74 & 45.59 \\
    ·Tent & 69.96 & 56.07 & 60.34 & 64.88 & 34.12 & 37.42 & 32.72 & 34.48 \\
    ·SAR  & 55.81 & 44.22 & 50.65 & 52.17 & 45.91 & 53.60 & 51.05 & 47.33 \\
    ·TLM  & 65.18 & \textbf{62.76} & 59.25 & 63.81 & 44.85 & 49.56 & 44.76 & 45.50 \\
    \rowcolor{salmon!20} ·NeFo & \textbf{75.26} & 59.77 & 63.25 & \textbf{69.76} & \textbf{52.78} & \textbf{57.90} & \textbf{51.31} & \textbf{53.39} \\
    \bottomrule
    \end{tabular}%

      \label{tab: TTL_comparison}%
\end{table*}%

\textbf{Online Adaptation.} \cref{tab: TTL_comparison} reports the performance on RS-Neg VQA and RS-Neg MCQ after online test-time adaptation. From the results, one could have the following observations: First, most existing TTA methods can further degrade model performance on negation understanding, which could be attributed to the self-reinforcement of wrong predictions. In contrast, NeFo could significantly improve the negation comprehension ability across diverse base models, e.g., it improves Qwen2.5-VL by 4.56\% and 11.17\% on RS-Neg VQA and RS-Neg MCQ, respectively. Second, across all negation types, NeFo achieves the most notable improvements on object-level queries, while gains on state-level queries are relatively modest, particularly on the VQA task (average 2.46\% versus 1.50\%). This discrepancy likely arises because object-level negation involves concrete, easily identifiable entities, whereas state-level negation requires reasoning about context-dependent conditions, which is inherently more challenging.

\begin{table*}[t]
  \centering
  \setlength{\tabcolsep}{2pt}
    \setlength{\aboverulesep}{0pt}
\setlength{\belowrulesep}{0pt}
\setlength{\extrarowheight}{0pt}
  \renewcommand\arraystretch{0.86}
      \caption{Zero-shot transfer evaluation on unseen datasets with negation queries. The best results are marked in \textbf{bold}.}
    \begin{tabular}{l|c|ccc|cc|c}
    \toprule
    \multirow{1}[4]{*}{Method} & Rs-Neg & \multicolumn{3}{c|}{RS-Neg Grounding} & \multicolumn{3}{c}{FloodNet VQA} \\
\cmidrule{3-8}          & Classification & IoU@0.25 & IoU@0.5 & IoU@0.75 & existence & counting & total \\
    \midrule
    Qwen2.5-VL & 60.37 & 72.54 & 54.03 & 27.66 & 40.14 & 35.23 & 37.32 \\
    ·Tent & 61.90 & 73.83 & 55.35 & 28.38 & 38.10 & 35.74 & 36.74 \\
    ·SAR  & 61.51 & 70.97 & 53.74 & 27.54 & 31.97 & 36.24 & 34.43 \\
    ·TLM  & 60.78 & 74.44 & 55.92 & 28.02 & 35.37 & 34.06 & 34.62 \\
    \rowcolor{salmon!20} ·NeFo & \textbf{66.63} & \textbf{81.16} & \textbf{63.37} & \textbf{34.46} & \textbf{82.54} & \textbf{38.93} & \textbf{57.47} \\
    \midrule
    Qwen3-VL & 51.31 & 82.65 & 69.00 & 41.10 & 94.78 & 39.26 & 62.87 \\
    ·Tent & 45.16 & 79.55 & 66.67 & 40.14 & 93.88 & 34.06 & 59.50 \\
    ·SAR  & 48.00 & 83.25 & 69.20 & \textbf{41.38} & 95.47 & 38.76 & 62.87 \\
    ·TLM  & 46.80 & 74.03 & 57.45 & 32.49 & 95.69 & 38.09 & 62.58 \\
    \rowcolor{salmon!20} ·NeFo & \textbf{54.75} & \textbf{83.49} & \textbf{69.24} & 40.94 & \textbf{96.15} & \textbf{40.10} & \textbf{63.74} \\
    \midrule
    RS-LLaVA & 44.18 & 22.75 & 3.06  & 0.04  & 94.10 & 20.13 & 52.36 \\
    ·Tent & 46.75 & 22.19 & 2.86  & 0.08  & 92.52 & 19.30 & 50.43 \\
    ·SAR  & 31.73 & 20.69 & 2.78  & 0.08  & 93.65 & 20.30 & 51.49 \\
    ·TLM  & 45.27 & 22.83 & 3.10  & 0.08  & 94.33 & 21.48 & 52.46 \\
    \rowcolor{salmon!20} ·NeFo & \textbf{48.47} & \textbf{25.89} & \textbf{4.23} & \textbf{0.28} & \textbf{95.24} & \textbf{22.32} & \textbf{53.33} \\
    \midrule
    GeoReason & 49.22 & \textbf{65.90} & \textbf{46.94} & \textbf{21.94} & 77.10 & 30.54 & 50.34 \\
    ·Tent & 41.59 & 4.31  & 1.97  & 0.60  & 70.75 & 23.49 & 43.59 \\
    ·SAR  & 39.98 & 13.37 & 6.48  & 2.05  & 78.46 & 28.19 & 49.57 \\
    ·TLM  & 50.20 & 27.90 & 16.63 & 5.43  & 78.46 & 28.02 & 49.47 \\
    \rowcolor{salmon!20} ·NeFo & \textbf{51.33} & 64.69 & 45.85 & \textbf{21.94} & \textbf{79.60} & \textbf{34.40} & \textbf{53.62} \\
    \bottomrule
    \end{tabular}%
      \label{tab: TTL_generalization}%
\end{table*}%

\textbf{Zero-shot Transferability.} To investigate whether NeFo can learn generalizable negation patterns, we evaluate the TTA-updated models on unseen negation data tasks. Specifically, we use the models after TTA on RS-Neg MCQ and test them on RS-Neg Classification, RS-Neg Visual Grounding, and FloodNet VQA \cite{rahnemoonfar2021floodnet}. Note that FloodNet is a flood disaster dataset, from which we curate 1,037 VQA samples with real-world negation queries, spanning existence and counting categories. As shown in \cref{tab: TTL_generalization}, for the scene-level classification task, NeFo consistently surpasses the original MLLMs, with improvements ranging from 2.11\% to 6.26\%. For the region-level grounding task, NeFo shows varied performance based on the original model. For example, it improves Qwen2.5-VL by 8.62\%, 9.34\%, and 6.80\% at IoU@0.25, IoU@0.5, and IoU@0.75, respectively. Notably, this performance even surpasses the results on corresponding affirmative queries (shown in \cref{fig:neg_vrs}), indicating that proper negation understanding provides additional discriminative information that helps the model better locate the target. In contrast, NeFo shows marginal performance drops on GeoReason (-1.21\% and -1.09\% at IoU@0.25 and IoU@0.5, respectively), but still far outperforms other TTA methods, e.g., TENT causes a catastrophic 61.59\% drop at IoU@0.25. This may be attributed to the reasoning-augmented nature of GeoReason, where answer-level supervision of TTA could disrupt the intermediate reasoning process. Our NeFo could mitigate this degradation through the knowledge retaining objective. For the real-world negation questions, i.e., FloodNet VQA, NeFo consistently achieves performance improvements, with an average gain of 2.63\% on the challenging counting task. These zero-shot results demonstrate that NeFo enables models to acquire generalizable negation understanding rather than simply fitting to the test data.

\subsection{Ablation Study}

\textbf{Impact of Each Component.} To study the influence of specific components in our method, we carry out the ablation study on the RS-Neg VQA and RS-Neg MCQ using Qwen2.5-VL-7B. Specifically, we ablate the contributions of three key training objectives of NeFo, i.e., Truth-Value Inversion (TI), Knowledge Retaining (KR), and Entropy Minimization (EM). From the results in \cref{tab: ablation_study}, we observe the following conclusions: First, the full NeFo could achieve the best performance, showing that all these objectives are important to improve the robustness against negation. Second, different tasks exhibit varying sensitivity to each component. For VQA, Truth-Value Inversion is most critical, whereas Knowledge Retaining plays a more important role in MCQ. This is likely because MCQ is more complex, containing both affirmative and negated captions, requiring the model to preserve its capability on affirmative queries to make correct distinctions.

\begin{table*}[t]
  \centering
  \setlength{\tabcolsep}{4pt}
    \setlength{\aboverulesep}{0pt}
\setlength{\belowrulesep}{0pt}
\setlength{\extrarowheight}{0pt}
  \renewcommand\arraystretch{0.86}
      \caption{Ablation study of each training objective on RS-Neg VQA and RS-Neg MCQ. }
    \begin{tabular}{l|ccc|c|ccc|c}
    \toprule
    \multirow{1}[4]{*}{Method} & \multicolumn{4}{c|}{RS-Neg VQA} & \multicolumn{4}{c}{RS-Neg MCQ} \\
\cmidrule{2-9}          & object & attribute & state & total & object & attribute & state & total \\
    \midrule
    Base  & 80.19 & 69.48 & 60.47 & 74.96 & 54.22 & 55.25 & 53.14 & 54.29 \\
    ·NeFo w/o TI & 79.22 & 68.89 & 59.17 & 74.05 & 57.97 & 58.28 & 54.45 & 57.78 \\
    ·NeFo w/o EM & 82.71 & 70.24 & 61.51 & 76.84 & 36.54 & 27.23 & 35.08 & 29.06 \\
    ·NeFo  w/o KR  & 83.81 & \textbf{72.68} & 58.91 & 77.82 & 20.32 & 31.48 & 32.72 & 22.71 \\
    \rowcolor{salmon!20} ·NeFo & \textbf{86.26} & 72.09 & \textbf{61.69} & \textbf{79.52} & \textbf{65.73} & \textbf{65.49} & \textbf{62.30} & \textbf{65.46} \\
    \bottomrule
    \end{tabular}%
      \label{tab: ablation_study}%
\end{table*}%

\begin{figure}[htbp]
    \centering
    \begin{subfigure}{0.49\textwidth}
        \centering
        \includegraphics[width=\textwidth]{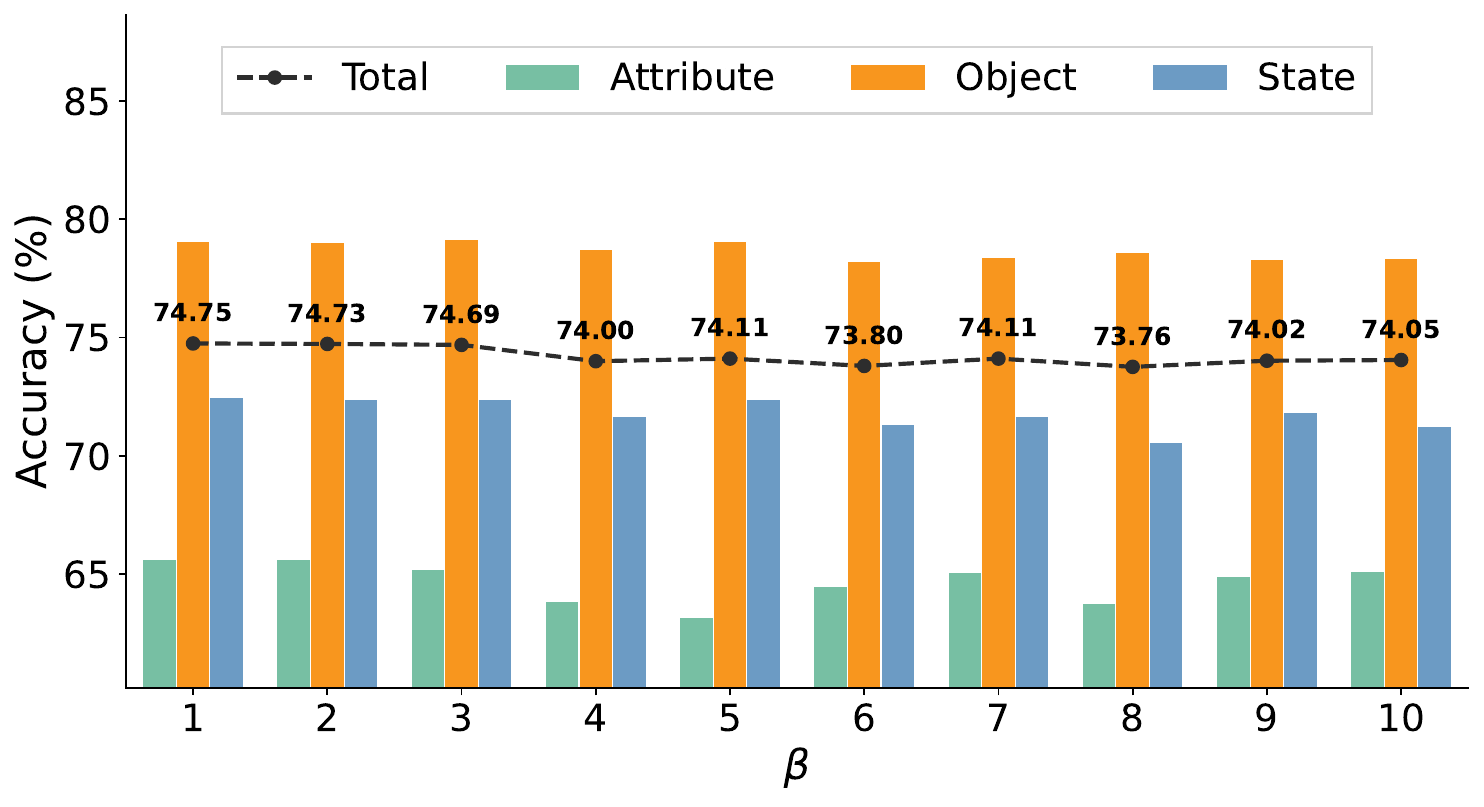}
        \caption{Effect of $\beta$ on RS-Neg VQA}
        \label{fig:pa_vqa_beta}
    \end{subfigure}
    \hfill
    \begin{subfigure}{0.49\textwidth}
        \centering
        \includegraphics[width=\textwidth]{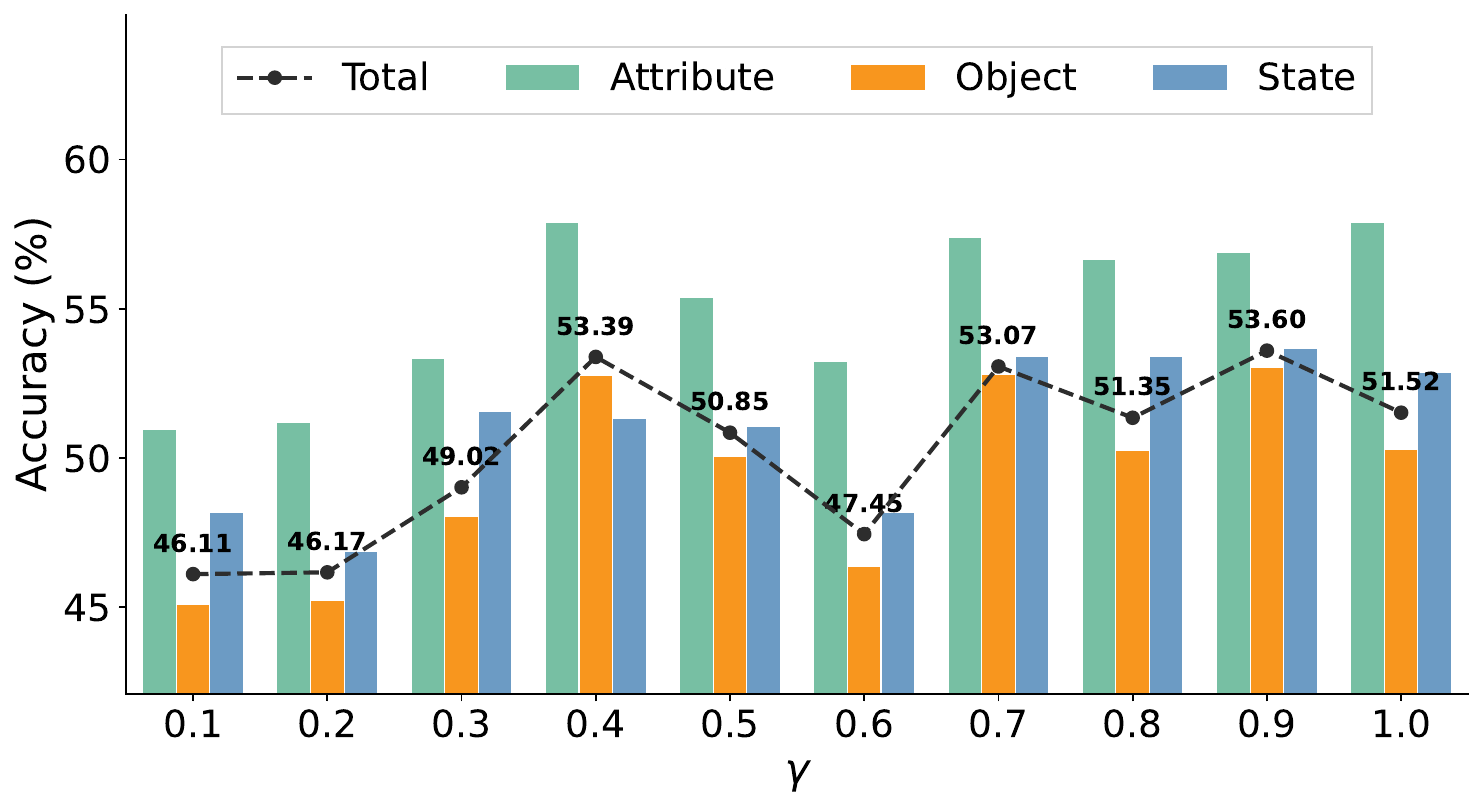}
        \caption{Effect of $\gamma$ on RS-Neg MCQ}
        \label{fig:pa_mcq_gamma}
    \end{subfigure}
    \caption{Sensitivity analysis of hyper-parameters in NeFo. (a) RS-Neg VQA with RS-LLaVA as the base model. (b) RS-Neg MCQ with GeoReason as the base model.}
    \label{fig:para_analysis}
\end{figure}

\textbf{Parameter Analysis.} We now investigate the effect of hyper-parameters $\beta$ and $\gamma$ by varying their values and plotting the resulting accuracy. Specifically, we report the accuracy on RS-Neg VQA with RS-LLaVA for $\beta$, and on RS-Neg MCQ with GeoReason for $\gamma$. As shown in \cref{fig:para_analysis}, we first observe that NeFo is robust to the choice of $\beta$, with VQA accuracy varying within a narrow range, i.e., $73.76\%\sim 74.75\%$. This is because a small $\beta$ is sufficient to prevent knowledge forgetting—the knowledge retaining loss consistently remains near zero during TTA, making further increases in $\beta$ unnecessary. In contrast, NeFo is more sensitive to $\gamma$, with performance dropping by 7.28\% when $\gamma$ is set to a small value ($\gamma$ = 0.1). This suggests that applying entropy minimization to negation-masked captions is necessary for the challenging MCQ task. On the one hand, it helps the model adapt to the domain distribution of test data. On the other hand, since $\mathcal{L}_{inv}$ relies on affirmative predictions as anchors, strengthening the model's confidence on affirmative inputs provides a more stable optimization direction for negation understanding. In addition, when $\gamma$ increases, NeFo can maintain stable performance across a broader range of values, i.e., $0.7 \sim 1.0$.

\begin{table*}[t]
  \centering
  \setlength{\tabcolsep}{4pt}
    \setlength{\aboverulesep}{0pt}
\setlength{\belowrulesep}{0pt}
\setlength{\extrarowheight}{0pt}
  \renewcommand\arraystretch{0.86}
      \caption{Comparisons of NeFo across different model sizes on RS-Neg VQA and RS-Neg MCQ. The best results are marked in \textbf{bold}.}
    \begin{tabular}{l|ccc|c|ccc|c}
    \toprule
    \multirow{1}[4]{*}{Method} & \multicolumn{4}{c|}{RS-Neg VQA} & \multicolumn{4}{c}{RS-Neg MCQ} \\
\cmidrule{2-9}          & object & attribute & state & total & object & attribute & state & total \\
    \midrule
    Qwen3-VL-2B & 70.29 & 55.82 & 54.74 & 64.61 & 34.74 & 42.48 & 40.31 & 36.19 \\
    \rowcolor{salmon!20} ·NeFo  & \textbf{82.06} & \textbf{62.42} & \textbf{62.29} & \textbf{74.51} & \textbf{37.69} & \textbf{45.39} & \textbf{42.67} & \textbf{39.10} \\
    \midrule
    Qwen3-VL-4B & 84.19 & 66.16 & 64.38 & 77.06 & 61.88 & 62.96 & 57.07 & 61.71 \\
    \rowcolor{salmon!20} ·NeFo  & \textbf{84.32} & \textbf{67.72} & \textbf{64.81} & \textbf{77.59} & \textbf{65.55} & \textbf{65.87} & \textbf{58.64} & \textbf{65.13} \\
    \midrule
    Qwen3-VL-8B & 76.00 & 63.64 & 65.33 & 71.47 & 57.29 & 58.15 & 55.76 & 57.30 \\
    \rowcolor{salmon!20} ·NeFo  & \textbf{81.04} & \textbf{66.08} & \textbf{66.99} & \textbf{75.42} & \textbf{64.90} & \textbf{63.84} & \textbf{61.78} & \textbf{64.55} \\
    \midrule
    Qwen3-VL-32B & 84.42 & 66.83 & 66.99 & 77.70 & 64.73 & 65.74 & 59.69 & 64.53 \\
    \rowcolor{salmon!20} ·NeFo  & \textbf{84.69} & \textbf{67.13} & \textbf{67.07} & \textbf{77.95} & \textbf{65.79} & \textbf{67.00} & \textbf{61.26} & \textbf{65.66} \\
    \bottomrule
    \end{tabular}%

      \label{tab: sacling_model}%
\end{table*}%

\subsection{Scaling Analysis}

\textbf{Scaling Across Model Sizes.} In this section, we explore the effectiveness of NeFo when applied to MLLMs of varying sizes. Specifically, we conduct experiments on the Qwen3-VL family across four model scales, i.e., 2B, 4B, 8B, and 32B. \cref{tab: sacling_model} presents the performance on RS-Neg VQA and RS-Neg MCQ. From the results, we could have the following observation: First, scaling up model size does not always improve performance, e.g., Qwen3-VL-4B significantly outperforms Qwen3-VL-8B on both VQA and MCQ tasks. Second, NeFo consistently improves performance across all four model scales. Notably, for the smallest 2B model, NeFo achieves a 9.9\% absolute gain on RS-Neg VQA, bringing its negation understanding capability close to that of the 32B model. This highlights the practical value of NeFo for resource-constrained deployments, such as edge devices or UAV-based RS applications.

\begin{figure}[t]
\centering  
\includegraphics[width=\columnwidth]{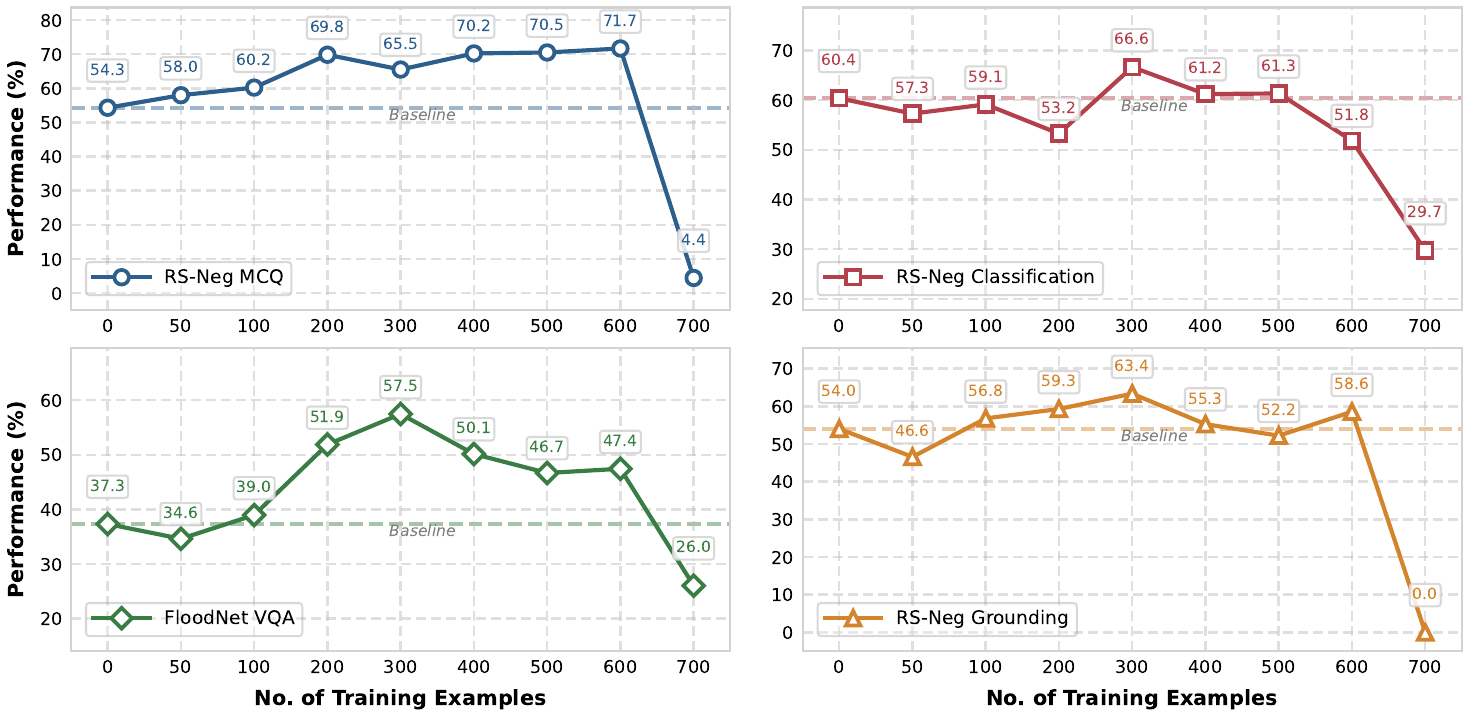}
\caption{Effect of NeFo with varying test-time adaptation data sizes. We report IoU@0.5 for RS-Neg Grounding and accuracy for the remaining tasks.}
\label{fig:scaling_data}
\end{figure}

\textbf{Scaling Across Data Sizes.} In this section, we study the effect of training data size on NeFo by varying the number of samples used for TTA. All experiments follow the same training settings for a fair comparison. We increase the training data for Qwen2.5-VL-7B on RS-Neg MCQ from 50 to 700 samples and evaluate its zero-shot performance on RS-Neg Classification, RS-Neg Grounding, and FloodNet VQA. As shown in \cref{fig:scaling_data}, one could see that model performance on MCQ improves significantly as training data increases, achieving a 17.4\% absolute gain with 600 training samples. However, further increasing the training data leads to overfitting and performance collapse, which may require a smaller learning rate to mitigate this issue. Meanwhile, the zero-shot performance on unseen datasets reveals a trade-off:  insufficient training samples cause the model to only fit the negation data without learning generalizable patterns, while excessive samples may lead to overfitting and degraded generalization.

\section{Conclusion}
This paper studies a practical but overlooked  problem in RS MLLMs: negation understanding. We introduce RS-Neg, a comprehensive benchmark covering four basic RS tasks, and reveal that current MLLMs suffer significant performance degradation under negation queries. To address this, we propose NeFo, a test-time adaptation method that incorporates negation logic into optimization via self-supervision. Experiments demonstrate that NeFo significantly improves negation understanding across different base models and generalizes well to unseen tasks. We hope this work could enhance the practicality of MLLMs in RS applications such as disaster response and emergency monitoring.

\section{Acknowledgmetns}
This work was supported in part by the Major Key Project of PCL under Grant PCL2025A10 and PCL2024A06, and in part by the Shenzhen Science and Technology Program under Grant RCJC20231211085918010.

%
%
\bibliographystyle{splncs04}
\bibliography{main}
\end{document}